\documentclass[runningheads]{llncs}
\usepackage{graphicx}
\usepackage{comment}
\usepackage{amsmath,amssymb} 

\usepackage[english]{babel}
\usepackage[T1]{fontenc}
\usepackage[utf8]{inputenc}
\usepackage{booktabs} 
\usepackage{soul}
\usepackage[titletoc]{appendix}

\makeatletter
\renewcommand*\env@matrix[1][\arraystretch]{%
  \edef\arraystretch{#1}%
  \hskip -\arraycolsep
  \let\@ifnextchar\new@ifnextchar
  \array{*\c@MaxMatrixCols c}}
\makeatother

\begin{document}
\pagestyle{headings}
\mainmatter
\def\ECCVSubNumber{3149}  

\title{Efficient Residue Number System Based Winograd Convolution} 

\titlerunning{Efficient Residue Number System Based Winograd Convolution}

\author{Zhi-Gang Liu \and Matthew Mattina} 
%
\authorrunning{Z.G. Liu et al.}

\institute{
Arm ML Research Lab, Boston, MA, USA 
\email{\{Zhi-Gang.Liu,Matthew.Mattina\}@arm.com} \\
}
\maketitle

\begin{abstract}
    Prior research has shown that Winograd algorithm can reduce the computational complexity of convolutional neural networks (CNN) with weights and activations represented in floating point. However it is difficult to apply the scheme to the inference of low-precision quantized (e.g. INT8) networks. Our work extends the Winograd algorithm to Residue Number System (RNS). The minimal complexity convolution is computed precisely over large transformation tile (e.g. $10 \times 10$ to $16 \times 16$) of filters and activation patches using the Winograd transformation and low cost (e.g. 8-bit) arithmetic without degrading the prediction accuracy of the networks during inference. The arithmetic complexity reduction is up to $7.03\times$ while the performance improvement is up to $2.30\times $ to $4.69\times $ for $3\times3$ and $5 \times 5$ filters respectively. 
   
\end{abstract}

\section{Introduction}

Machine learning has achieved great success in the past decade on a variety of applications including computer vision, natural language processing, and automatic speech recognition. In particular, deep convolutional neural networks (CNNs) have achieved better than human-level accuracy on image classification. The learning capability of CNNs improves with increasing depth and number of channels in the network layers. However this improvement comes at the expense of growing computation cost, particularly the expensive matrix or tensor multiplication and convolution. Thus reducing the computational complexity, especially the cost of the convolution operations, is critical for the deployment of these models on mobile and embedded devices with limited processing power. 

Most recent CNN architectures~\cite{CNN} for image classification use low dimensional filters, typically 3$\times$3, 5$\times$5 or 7$\times$7. The conventional Fast Fourier Transform (FFT) based convolution in the complex domain is inefficient with small filter dimensions. Faster algorithms for CNN inference based on Winograd minimal filters~\cite{shumel-winograd} can speed up the convolution by a factor of 2 to 4. The downside of the Winograd approach is that numerical problems and accuracy loss can occur unless high precision floating-point values are used.

Research on the quantization of neural network~\cite{binary2}~\cite{LM} has shown that using reduced-precision representation (e.g. INT8) for the storage and computation of CNNs has significant benefits such as decreased memory bandwidth, lower memory foot-print, lower power consumption and higher throughput, while only having a negligible prediction accuracy degradation. The predominant numerical format used for training neural networks is IEEE floating-point format (FP32). There is a potential $4\times$ reduction in memory bandwidth and storage achieved by quantizing FP32 weights and activations to INT8 values. The corresponding energy and area saving are 13.5$\times$ and 27.3$\times$~\cite{dally} respectively. But, both Winograd and FFT methods~\cite{winograd}~\cite{fft} require high precision arithmetic to avoid prediction accuracy degradation and are therefore non-ideal for improving low-precision integer e.g. INT8 convolution efficiently.

In this paper, we extend the Winograd minimal convolution~\cite{shumel-winograd} to Residue Number System (RNS) ~\cite{RNS} targeting the inference of low-precision e.g. INT8 quantized convolutional neural networks. The key contributions are summarized here:
\begin{itemize}
\item {
We formulate the Winograd minimal complexity integer convolution over Residue Number System (RNS). The use of the RNS enables our algorithm to operate on quantized, low-precision e.g. INT8 CNNs with low cost, low precision integer arithmetic, without computational instability issues and without impacting the accuracy of the networks.  }
\item{
Our RNS-based formulation enables the use of much larger Winograd transformation tiles, e.g. from 8x8 to 16x16. The theoretical arithmetic reduction is up to 2.3$\times$ and 4.69$\times$ for $3\times3$ and $5\times5$ filters respectively over 3-residue power efficient 8-bit RNS; 3.45$\times$ and $7.03\times$ for 2-residue 16-bit RNS.} 
\item{
We analyzed the performance with 8-bit quantized VGG16 models and show 2.02$\times$ to 2.2$\times$ improvement of inference latency on Arm Cortex-A73 CPU.}
\end{itemize}

\section{Related Work}
Earlier work applied the classical FFT to speedup convolutional layers by reducing the arithmetic complexity~\cite{fft}. This approach requires float arithmetic in the complex number $\textbf{C}$, and multiplication involves the real and imaginary parts of complex value. A product of two complex values needs 3 or 4 floating multiplications, which is inefficient, especially for the small size filters commonly defined in popular CNNs.

The Winograd minimal filtering algorithm~\cite{shumel-winograd}, first applied to CNNs by Lavin and Gray~\cite{winograd}, can reduce arithmetic complexity from 2.25$\times$ to 4$\times$ for typical $3\times3$ CNN filters. However, the algorithm requires high precision arithmetic and hits computational instability issues when applied to large transform tile sizes~\cite{winograd_error}. An efficient sparse implementation of Winograd convolution have also been proposed~\cite{sparse_winograd}. The conventional Winograd convolution algorithm, including the latest enhancements, requires high precision floating point arithmetic.

Meanwhile, some researchers have tried to extend the Winograd algorithm to reduced-precision integer arithmetic by choosing complex interpolation points~\cite{mb} with a 17.37\% throughput improvement claimed, however it depends on a lossy precision scaling scheme, which would cause predication accuracy drop. 

\section{Residue Number System ($\textbf{RNS}$)} \label{sec:rns}
A Residue Number System, RNS$(m_0, m_1, .., m_{n-1})$~\cite{RNS}, is number system to represent an integer by its value modulo $n$ pairwise coprime moduli $ m_0, m_1, .., $ and $ m_{n-1}$. 
\begin{equation}\nonumber
\begin{split}
    &x_0 = x \pmod {m_0}   \\
    &x_1 = x \pmod {m_1}   \\
    &\ \ \ ... \\
    &x_{n-1} = x \pmod {m_{n-1}}
\end{split}
\end{equation}
For example, to represent $x=48$ in RNS$(m_0=7, m_1=9)$    
\begin{equation} \nonumber
\begin{split}
\{x\pmod{m_0}, x\pmod{m_1}\}= \{6, 3\} \\
\end{split}
\end{equation}

We can construct the value of $x$ from its RNS representation as long as $x < M$, where $M = {\displaystyle \prod_{i=1}^{n-1}}m_i$ is the dynamic range of the RNS$(m_0,m_1,..,m_{n-1})$. For example, to convert $\{ 6, 3\}$ from RNS(7,9) back to standard form using Mixed Radix Conversion(MRC) \cite{MRC} or Chinese Remainder Theorem (CRT) ~\cite{knuth_CRT}.
\begin{equation} \nonumber
\begin{split}
    x = \big[6 + 7*[\frac{3-6}{7}\pmod{9}]\big] \pmod{7*9} = 6 + 7*6 = 48
\end{split}
\end{equation}

For addition($+$), subtraction($-$) and multiplication($*$) of two RNS values $ x=\{x_0, x_1, .., x_{n-1}\} $ and $ y=\{y_0, y_1, .., y_{n-1}\}$, it's sufficient to perform the  operation on corresponding pair of residues. For example, $x =\{6, 3\}, y = \{5, 10\} \in$ RNS(7,9)
\begin{equation}\nonumber
\begin{split}
&x + y = \{6 + 5 \pmod{7}, 3 + 10 \pmod{9}\} = \{ 4, 4\} \\
&x - y = \{6 - 5 \pmod{7}, 3 - 10 \pmod{9}\} = \{ 1, 2\} \\
&x *\  y = \{6 *\  5 \pmod{7}, 3\ * 10 \pmod{9}\} = \{ 2, 3\} \\
\end{split}
\end{equation}

\subsection{Convolution in $\textbf{RNS}$} \label{sec:conv_rns}

Equivalently, we could calculate the convolution $y$ of $N$-element vector $\textbf{\textit{d}} = (d_0, d_1, d_2, .., d_{N-1})$ and $R$-element filter $\textbf{\textit{g}} = ( g_0, g_1, g_2, .., g_{R-1} )$  over RNS$(m_0, $ $ m_1,$ $ .., m_{n-1})$.
\begin{equation} \label{eq:conv} \nonumber 
\textbf{\textit{y}}  = (y_0, y_1, y_2, .., y_{N-R}) = \textbf{\textit{d}} \circledast \textbf{\textit{g}} \\
\end{equation}
and $  y_k = \{y_{k}^{(0)}, y_{k}^{(1)}, .., y_{k}^{(n-1)}\} \in$ RNS$(m_0, m_1,.., m_{n-1}) $, where
\begin{equation} \label{eq:conv_rns} \nonumber 
 y_{k}^{(i)} = ({\sum_{j=0}^{R-1}{ d_{k+j} * g_j }}) \pmod{m_i}
\end{equation}


\section{Winograd Convolution} \label{sec:winograd} 
The Winograd convolution~\cite{shumel-winograd} is an optimal algorithm to compute short convolution over real numbers, outperforming conventional Discrete Fourier Transform (DFT). 
$F(M, R)$ denotes the convolution computation of $M$-tuple output $y$ of a $R$-tuple filter $g$ and $N$-tuple input $d$ where $N=M + R -1$. The Winograd algorithm calculates the $F(M, R)$ in a bilinear form as

\begin{equation} \nonumber 
       y = A^{T} \Big[(G g) \odot (B^T d)\Big]
\end{equation}
where $\odot$ acts as element-wise production and 
$B^T$, $G$ and $A^{T}$ are  $N \times N$, $N \times R$ and $M \times N$ transform matrices respectively. 

Specifically, $A^T, G $ and $B^T$ are derived from the Vandermonde matrix \footnote{\url{https://en.wikipedia.org/wiki/Vandermonde\_matrix}} $V$ generated from $N$ distinct Lagrange interpolation points $S_0, S_1, S_2, .. S_{N-1} $ (Note: Require a special handling if $S_{N-1} = \infty$).  
\begin{equation} \label{eq:vandermonde} V = 
    \begin{pmatrix}
    1 \ \  &  S_0            & S^2_0             &\ ...   & \ \ \ S_0^{N-1}       \\
    1 \ \  &  S_1            & S^2_1             &\ ...   & \ \ \ S_1^{N-1}       \\
    1 \ \  &  S_2            & S^2_2             &\ ...   & \ \ \ S_2^{N-1}       \\
    ..     &  ..             &  ..               &\ ...   & ..                    \\
    1 \ \  & \ \ \ S_{{N-1}} & \ \ \ S_{{N-1}}^2 &\ ...   & \ \ \ S_{{N-1}}^{N-1} \\
    \end{pmatrix}_{_{N\times N}}  
\end{equation}
\normalsize 
and
\begin{align} \nonumber
\scriptsize
    &A^T = V^T_{ \ \ [0:M-1;0:N-1]}  \\
    \nonumber
    &G = V_{\ \ [0:N-1;0:R-1]}  \\
    \nonumber
    &B^T = V^{-T}    
\end{align}
\normalsize
For 2-D convolution, similar fast algorithm $F(M \times M, R \times R) $ can be represented as 
\begin{equation} \label{eq:fast_algorithm}   
       {y} = A^{T} \Big[(G {g} G^T) \odot (B^T {d} B)\Big] A
\end{equation}
We call $ G g G^T$ and $ B^T d B$ the $\textit{forward transform}$ and $A^T[\cdot]A$ the $\textit{backward transform}$. 

Assuming the computation cost of transformation $G g G^T$ and $ B^T d B $ was amortized completely due to reuse, the fast algorithm requires $N ^ 2=(M+R-1)^2$ multiplications while the standard method uses $M^2 R^2$. The arithmetic complexity reduction is
$\frac{M^2 R^2}{{(M + R - 1)}^2}$.
For example:

$F(2 \times 2, 3 \times 3)$ with interpolation points $\{ 0, \pm 1, \infty \}$. The fractions in $B^T$ are arranged into matrix $G$. The arithmetic complexity reduction is $2.25\times$. 
\scriptsize 
\begin{align*}
& A^T = \begin{pmatrix}
1 & 1 & 1 & 0 \\
0 & 1 & -1 & 1
\end{pmatrix}; \ \
B^T =\begin{pmatrix}
1 & 0 & -1 & 0 \\
0 & 1 & 1 & 0 \\
0 & -1 & 1 & 0 \\
0 & -1 & 0 & 1
\end{pmatrix}; \ \
G  = \begin{pmatrix}[1.2]
1 & 0 & 0 \\
\frac{1}{2}  & \frac{1}{2}  & \frac{1}{2}  \\
\frac{1}{2}  & \frac{-1}{2}  & \frac{1}{2}  \\
0 & 0 & 1
\end{pmatrix} = \frac{1}{2} G^{'} ; \ \ G^{'} = \begin{pmatrix}
2 & 0 & 0 \\
1 & 1 & 1 \\
1 & -1 & 1\\
0 & 0 & 2
\end{pmatrix} \\
\end{align*}
\normalsize

$F(4 \times 4, 3 \times 3)$ with interpolation points $\{ 0, \pm1, \pm2, \infty \}$. The arithmetic complexity reduction is $4\times$.
\scriptsize 
\tiny
\begin{align*} 
& A^T \! = \! \begin{pmatrix}
1 & 1 & 1 & 1 & 1 & 0 \\
0 & 1 & -1 & 2 & -2 & 0 \\
0 & 1 & 1 & 4 & 4 & 0 \\
0 & 1 & -1 & 8 & -8 & 1
\end{pmatrix};\   
B^T \! = \! \begin{pmatrix}
4 & 0 & -5 & 0 & 1 & 0 \\
0 & 4 & 4 & -1 & -1 & 0 \\
0 & -4 & 4 & 1 & -1 & 0 \\
0 & -2 & -1 & 2 & 1 & 0 \\
0 & 2 & -1 & -2 & 1 & 0 \\
0 & 4 & 0 & -5 & 0 & 1
\end{pmatrix};\ 
G \! = \! \begin{pmatrix} [1.5]
\frac{1}{4}  & 0 & 0 \\
\frac{1}{6}  & \frac{1}{6}  & \frac{1}{6}  \\
\frac{1}{6}  & \frac{-1}{6}  & \frac{1}{6}  \\
\frac{1}{24}  & \frac{1}{12}  & \frac{1}{6}  \\
\frac{1}{24}  & \frac{-1}{12}  & \frac{1}{6}  \\
0 & 0 & 1
\end{pmatrix} \! = \! \frac{1}{24}G^{'};\ 
G^{'} \!=\!
\begin{pmatrix} 
6 & 0 & 0 \\
4  & 4  & 4  \\
4  & -4 & 4  \\
1  & 2  & 4  \\
1  & -2 & 4  \\
0 & 0 & 24
\end{pmatrix} \\
\end{align*}
\normalsize 
where matrices $A^T, G$ and $B^T$ are derived from Vandermonde matrix of the roots to construct the transform. 

$F(2 \times 2, 3 \times 3)$ and $F(4 \times 4, 3 \times 3)$ have theoretical arithmetic complexity reduction of $2.25\times$ and $4\times$ respectively. We can achieve the expected speedup using floating-point operation i.e. FP32. However, it's a challenge to implement the Winograd convolution using low-precision integral arithmetic for quantized CNN.        
To calculate exact convolution using integer arithmetic, we can obtain matrix $G^{'}$ by factoring out the common fraction $\alpha$, e.g. $\alpha = \frac{1}{2}$ for $F(2 \times 2, 3 \times 3)$ and $\alpha=\frac{1}{24}$ for $F(4 \times 4, 3 \times 3)$, from corresponding matrix $G$. Then eq. \ref{eq:fast_algorithm} becomes $ {y} = \alpha^2 A^{T} \Big[(G^{'} {g} {G^{'}}^T) \odot (B^T {d} B)\Big] $. 

The magnitude of element in transformation $G^{'} g {G^{'}}^T$ and $B^T d B $ would be $\frac{trace({G^{'} {G^{'}}^T})}{N} $ and $\frac{trace({B^T B})}{N}$ times as large as the quantity of filter $g$ and input $d$ on average. Particularly, the magnification are $3.5\times $ and $2 \times$ for $F(2 \times 2, 3 \times 3)$ and $125\times $ and $28.7 \times$ for $F(4 \times 4, 3 \times 3)$. Moreover, the magnifications we calculated correspond to the standard deviation statistically, the outliers could have much larger magnitudes. Practically, we need 12 bits to hold each element of transformation and INT16 arithmetic for element-wise multiply for $F(2 \times 2, 3 \times 3)$. $F(4 \times 4, 4 \times 3)$ demands 18 bits for transformation and INT32 arithmetic operations. We summarized the data width of transformation and arithmetic reduction of integer Winograd algorithms in table ~\ref{tab:operations}. Although the Winograd algorithms enable complexity reduction, they require higher precision arithmetic than INT8. 
\begin{table*}[bt] 
\caption{The required data width of transformation and the corresponding arithmetic reduction for integer (INT8) Wingograd convolution algorithms. $DW$ is transformation data width in bit. $Arithmetic\ Reduction$ is the reduction of operation in $DW$ bits.} 
\label{tab:operations}
\scriptsize 
\centering
\begin{tabular}{l | c | c  }
\toprule
Algorithm \ \  & \ \ DW (bit) \ \  & \ \ Arithmetic Reduction\ \  \\
\midrule
$F$($2\times2$, $3\times 3$)   & 12 & 2.25$\times$ \\
$F$($4\times4$, $3\times 3$)   & 18 & 4.00$\times$ \\
$F$($6\times6$, $3\times 3$)   & 24 & 5.06$\times$ \\
$F$($8\times8$, $3\times 3$)   & 36 & 5.76$\times$ \\
$F$($8\times8$, $5\times 5$)   & 43 & 11.1$\times$ \\
$F$($10\times10$, $3\times 3$) \  & 50 & 6.26$\times$  \\ 
$F$($10\times10$, $5\times 5$)    & 60 & 12.7$\times$  \\ 
\bottomrule 
\end{tabular} \\ 
\end{table*}
Considering INT8 multipliers cost about $\frac{1}{4}$ power and area of INT16 case; $\frac{1}{15}$ and $\frac{1}{12}$ of INT32; $\frac{1}{18}$ and $\frac{1}{27}$ of FP32 respectively~\cite{dally}, there will be advantage in implementing the Winograd algorithm using INT8 arithmetic. For this reason, a lossy precision scaling scheme was proposed~\cite{mb}, which scales down the transformation in the range of the desired low-precision arithmetic operation. However the scaling method introduces errors to the convolution output and would cause predication accuracy degradation.       

The fundamental difficulty with performing the standard Winograd algorithm using low cost integral arithmetic is due to the ill-conditioned Vandermonde (and its inverse) matrix $V$ in eq. ~\ref{eq:vandermonde} with real interpolation points especially for large transformation (e.g. M > 6). 
We propose a different approach to implement the Winograd algorithm over Residue Number System (RNS) via low-precision integer arithmetic (e.g. INT8 or INT16) in the next section.   

\section{Winograd Convolution over Residue Number System}  \label{sec:winograd_rns}
We extend the Winograd convolution algorithm described in section ~\ref{sec:winograd} to Residue Number System (RNS) in section ~\ref{sec:rns} to formulate a new implementation. This new approach solves the numerical stability issue of the conventional Winograd algorithm for large transformation, i.e. $ M \in [8, 16] $, moreover the new method is compatible with low precision 8-bit multiply and accumulation.

To simplify the description, without loss of generality, we take $F(10 \times 10, 3 \times 3)$ with interpolation points $\{0, \pm1, \pm2, \pm3, \pm4, \pm5,  \infty \}$ as a running example with the following transform matrices $A^T, B^T$ and $G$.  

\tiny 
\setcounter{MaxMatrixCols}{20}
\begin{align*}
& A^T = \begin{pmatrix}
1 & 1 & 1 & 1 & 1 & 1 & 1 & 1 & 1 & 1 & 1 & 0 \\
0 & 1 & -1 & 2 & -2 & 3 & -3 & 4 & -4 & 5 & -5 & 0 \\
0 & 1 & 1 & 4 & 4 & 9 & 9 & 16 & 16 & 25 & 25 & 0 \\
0 & 1 & -1 & 8 & -8 & 27 & -27 & 64 & -64 & 125 & -125 & 0 \\
0 & 1 & 1 & 16 & 16 & 81 & 81 & 256 & 256 & 625 & 625 & 0 \\
0 & 1 & -1 & 32 & -32 & 243 & -243 & 1024 & -1024 & 3125 & -3125 & 0 \\
0 & 1 & 1 & 64 & 64 & 729 & 729 & 4096 & 4096 & 15625 & 15625 & 0 \\
0 & 1 & -1 & 128 & -128 & 2187 & -2187 & 16384 & -16384 & 78125 & -78125 & 0 \\
0 & 1 & 1 & 256 & 256 & 6561 & 6561 & 65536 & 65536 & 390625 & 390625 & 0 \\
0 & 1 & -1 & 512 & -512 & 19683 & -19683 & 262144 & -262144 & 1953125 & -1953125 & 1
\end{pmatrix} \\
& B^T =\begin{pmatrix}
14400 & 0 & -21076 & 0 & 7645 & 0 & -1023 & 0 & 55 & 0 & -1 & 0 \\
0 & 14400 & 14400 & -6676 & -6676 & 969 & 969 & -54 & -54 & 1 & 1 & 0 \\
0 & -14400 & 14400 & 6676 & -6676 & -969 & 969 & 54 & -54 & -1 & 1 & 0 \\
0 & -7200 & -3600 & 8738 & 4369 & -1638 & -819 & 102 & 51 & -2 & -1 & 0 \\
0 & 7200 & -3600 & -8738 & 4369 & 1638 & -819 & -102 & 51 & 2 & -1 & 0 \\
0 & 4800 & 1600 & -6492 & -2164 & 1827 & 609 & -138 & -46 & 3 & 1 & 0 \\
0 & -4800 & 1600 & 6492 & -2164 & -1827 & 609 & 138 & -46 & -3 & 1 & 0 \\
0 & -3600 & -900 & 5044 & 1261 & -1596 & -399 & 156 & 39 & -4 & -1 & 0 \\
0 & 3600 & -900 & -5044 & 1261 & 1596 & -399 & -156 & 39 & 4 & -1 & 0 \\
0 & 2880 & 576 & -4100 & -820 & 1365 & 273 & -150 & -30 & 5 & 1 & 0 \\
0 & -2880 & 576 & 4100 & -820 & -1365 & 273 & 150 & -30 & -5 & 1 & 0 \\
0 & -14400 & 0 & 21076 & 0 & -7645 & 0 & 1023 & 0 & -55 & 0 & 1
\end{pmatrix}
\end{align*}
\begin{align*}
& G  = \begin{pmatrix} [1.8]
\frac{1}{14400}  & 0 & 0 \\
\frac{1}{17280}  & \frac{1}{17280}  & \frac{1}{17280}  \\
\frac{1}{17280}  & \frac{-1}{17280}  & \frac{1}{17280}  \\
\frac{1}{30240}  & \frac{1}{15120}  & \frac{1}{7560}  \\
\frac{1}{30240}  & \frac{-1}{15120}  & \frac{1}{7560}  \\
\frac{1}{80640}  & \frac{1}{26880}  & \frac{1}{8960}  \\
\frac{1}{80640}  & \frac{-1}{26880}  & \frac{1}{8960}  \\
\frac{1}{362880}  & \frac{1}{90720}  & \frac{1}{22680}  \\
\frac{1}{362880}  & \frac{-1}{90720}  & \frac{1}{22680}  \\
\frac{1}{3628800}  & \frac{1}{725760}  & \frac{1}{145152}  \\
\frac{1}{3628800}  & \frac{-1}{725760}  & \frac{1}{145152}  \\
0 & 0 & 1
\end{pmatrix} = \frac{1}{3628800} G^{'}; \ \ \ \ 
G^{'} = \begin{pmatrix} 
252 & 0 & 0 \\
210  & 210  & 210  \\
210 & -210  & 210  \\
120  & 240  & 480  \\
120  & -240  & 480  \\
45  & 135 & 405  \\
45  & -135  & 405  \\
10  & 40  & 160  \\
10  & -40  & 160  \\
1  & 5  & 25  \\
1  & -5  & 25  \\
0 & 0 & 3628800
\end{pmatrix} 
\end{align*}
\normalsize

These transforms are derived from the $12 \times 12$ Vandermonde matrix and its inverse matrix \footnote{\label{note1}\url{ https://proofwiki.org/wiki/Inverse\_of\_Vandermonde\_Matrix}}, which are not computationally friendly in standard number systems, including FP32 due to its numerical instability. 
However, we could mitigate the instability by carrying out the computation of eq. \ref{eq:fast_algorithm} over RNS$(m_0, m_1, .., m_{n-1})$.

To represent the transform matrix $G$ in RNS, the modulus $m_0, m_1, ..,$ and $m_{n-1}$ need be coprime to $\frac{1}{\alpha}$, e.g. $\frac{1}{\alpha} = 3628800 = 2^8 \cdot 3^4 \cdot 5^2 \cdot 7$ for the $F(10\times 10,3 \times 3)$ example.

Generically, the inverse of $N \times N$ Vandermonde matrix $V$ in eq. ~\ref{eq:vandermonde}  \textsuperscript{\ref{note1}}~\cite{knuth}, $V^{-1} = \{V^{-1}_{i,j}\}$, and $i, j \in [0,N-1]$ and $V^{-1}_{i,j}$ is given in eq. \ref{eq:inverse_vandermonde}.   
\begin{equation}  \label{eq:inverse_vandermonde}
    V^{-1}_{i,j} = 
    \begin{cases}
        \frac{1}{\prod\limits_{m=0, \ m \ne j}^{N-1}{(S_j - S_m)}} & for \ j = N-1 \\
        \\
        \frac{(-1)^{N-1-i}\sum\limits_{0 \le j_0 < j_1 < ... < j_{_{N-1-i}} < N, \ j_k \ne j}{S_{j_0}S_{j_1}...S_{j_{_{N-1-i}}}     }}{\prod\limits_{m=0, m \ne j}^{N-1}{(S_j - S_m)}} &  otherwise \\
    \end{cases} 
\end{equation} 
where $S_0, S_1, S_2,...,S_{N-1}$ are the interpolation points we choose to construct the Winograd transform. To obtain the multiplicative inverse of the denominator of $V^{-1}_{i,j}$ in eq. ~\ref{eq:inverse_vandermonde}, each modulus $m_i$ need be coprime to the denominator $\prod\limits_{m=0, \ m \ne j}^{N-1}{(S_j - S_m)}$. 

For our example, the denominators in $G$ are $14400=2^6 \cdot 3^2 \cdot 5^2$,  $17280=2^7 \cdot 3^3 \cdot 5$, $30240=2^5 \cdot 3^5 \cdot 5 \cdot 7$, $80640=2^8 \cdot 3^2 \cdot 5 \cdot 7$, $362880=2^7 \cdot 3^4 \cdot 5 \cdot 7$ and $3628800=2^8 \cdot 3^4 \cdot 5^2 \cdot 7$.   
We chose moduli $m_0 = 11 \times 23$ = 253, $m_1=251$ and $m_2=13 \times 19$ = 247, which are all coprime to the denominators in $G$. Therefore the fractions in matrix $G$ are all well-defined for modular division, for instance $\frac{1}{14400} \pmod{253} = 12$ as a result of multiplicative inverse of denominator, e.g. $14400 \times 12 \pmod{253} = 1$. Similarly,  $\frac{1}{14400} \pmod{251} = 27$ and $\frac{1}{14400} \pmod{247} = -10$. Moreover, moduli $(253, 251, 247)$ are the largest suitable 8-bit values for the interpolation points we chose. Given that we can convert matrix $A^T, G$ and $B^T$ to corresponding modular format,  e.g. $A^T_{m_{i}} = A^T \pmod{m_i}$,  $G_{m_{i}} = G \pmod{m_i}$ and  $B^T_{m_{i}} = B^T \pmod{m_i}$, where $m_i \in (253, 251, 247)$. The RNS representation of eq. \ref{eq:fast_algorithm} is
\begin{align} \label{eq:fast_algorithm_rns}
y =
(A^T_{_{253}} \Big[[G_{_{253}} {g} G^T_{_{253}}] \odot [B^{T}_{_{253}}{d} B_{_{253}}]\Big] A_{_{253}},\nonumber\\
A^T_{_{251}} \Big[[G_{_{251}} {g} G^T_{_{251}}] \odot [B^{T}_{_{251}}{d} B_{_{251}}]\Big] A_{_{251}},\nonumber\\
A^T_{_{247}} \Big[[G_{_{247}} {g} G^T_{_{247}}] \odot [B^{T}_{_{247}}{d} B_{_{247}}]\Big] A_{_{247}})
\end{align}

For modulo 253, the corresponding transform matrices are 
\scriptsize 
\tiny
\begin{align*}
& {A^T_{_{253}}} = \begin{pmatrix}
1 & 1 & 1 & 1 & 1 & 1 & 1 & 1 & 1 & 1 & 1 & 0 \\
0 & 1 & -1 & 2 & -2 & 3 & -3 & 4 & -4 & 5 & -5 & 0 \\
0 & 1 & 1 & 4 & 4 & 9 & 9 & 16 & 16 & 25 & 25 & 0 \\
0 & 1 & -1 & 8 & -8 & 27 & -27 & 64 & -64 & 125 & -125 & 0 \\
0 & 1 & 1 & 16 & 16 & 81 & 81 & 3 & 3 & 119 & 119 & 0 \\
0 & 1 & -1 & 32 & -32 & -10 & 10 & 12 & -12 & 89 & -89 & 0 \\
0 & 1 & 1 & 64 & 64 & -30 & -30 & 48 & 48 & -61 & -61 & 0 \\
0 & 1 & -1 & -125 & 125 & -90 & 90 & -61 & 61 & -52 & 52 & 0 \\
0 & 1 & 1 & 3 & 3 & -17 & -17 & 9 & 9 & -7 & -7 & 0 \\
0 & 1 & -1 & 6 & -6 & -51 & 51 & 36 & -36 & -35 & 35 & 1
\end{pmatrix}; \ \ 
G_{_{253}}  = \begin{pmatrix}
12 & 0 & 0 \\
10 & 10 & 10 \\
10 & -10 & 10 \\
78 & -97 & 59 \\
78 & 97 & 59 \\
-34 & -102 & -53 \\
-34 & 102 & -53 \\
-120 & 26 & 104 \\
-120 & -26 & 104 \\
-12 & -60 & -47 \\
-12 & 60 & -47 \\
0 & 0 & 1
\end{pmatrix}\\ 
&{B^T_{_{253}}} = \begin{pmatrix}
-21 & 0 & -77 & 0 & 55 & 0 & -11 & 0 & 55 & 0 & -1 & 0 \\
0 & -21 & -21 & -98 & -98 & -43 & -43 & -54 & -54 & 1 & 1 & 0 \\
0 & 21 & -21 & 98 & -98 & 43 & -43 & 54 & -54 & -1 & 1 & 0 \\
0 & -116 & -58 & -117 & 68 & -120 & -60 & 102 & 51 & -2 & -1 & 0 \\
0 & 116 & -58 & 117 & 68 & 120 & -60 & -102 & 51 & 2 & -1 & 0 \\
0 & -7 & 82 & 86 & 113 & 56 & 103 & 115 & -46 & 3 & 1 & 0 \\
0 & 7 & 82 & -86 & 113 & -56 & 103 & -115 & -46 & -3 & 1 & 0 \\
0 & -58 & 112 & -16 & -4 & -78 & 107 & -97 & 39 & -4 & -1 & 0 \\
0 & 58 & 112 & 16 & -4 & 78 & 107 & 97 & 39 & 4 & -1 & 0 \\
0 & 97 & 70 & -52 & -61 & 100 & 20 & 103 & -30 & 5 & 1 & 0 \\
0 & -97 & 70 & 52 & -61 & -100 & 20 & -103 & -30 & -5 & 1 & 0 \\
0 & 21 & 0 & 77 & 0 & -55 & 0 & 11 & 0 & -55 & 0 & 1
\end{pmatrix}
\end{align*}
\normalsize 
All elements in these matrices are in the range of $[-\frac{253-1}{2}, \frac{253-1}{2}]$. The computation of the fast convolution over $\pmod{253}$ can be performed with 8-bit low cost arithmetic operation without numerical concerns. Similarly, we can get the transforms for 251 and 247.   
\scriptsize 
\tiny
\begin{align*}
& A^T_{_{251}} = \begin{pmatrix}
1 & 1 & 1 & 1 & 1 & 1 & 1 & 1 & 1 & 1 & 1 & 0 \\
0 & 1 & -1 & 2 & -2 & 3 & -3 & 4 & -4 & 5 & -5 & 0 \\
0 & 1 & 1 & 4 & 4 & 9 & 9 & 16 & 16 & 25 & 25 & 0 \\
0 & 1 & -1 & 8 & -8 & 27 & -27 & 64 & -64 & 125 & -125 & 0 \\
0 & 1 & 1 & 16 & 16 & 81 & 81 & 5 & 5 & 123 & 123 & 0 \\
0 & 1 & -1 & 32 & -32 & -8 & 8 & 20 & -20 & 113 & -113 & 0 \\
0 & 1 & 1 & 64 & 64 & -24 & -24 & 80 & 80 & 63 & 63 & 0 \\
0 & 1 & -1 & -123 & 123 & -72 & 72 & 69 & -69 & 64 & -64 & 0 \\
0 & 1 & 1 & 5 & 5 & 35 & 35 & 25 & 25 & 69 & 69 & 0 \\
0 & 1 & -1 & 10 & -10 & 105 & -105 & 100 & -100 & 94 & -94 & 1
\end{pmatrix};\ \ 
G_{_{251}}  = \begin{pmatrix}
27 & 0 & 0 \\
-103 & -103 & -103 \\
-103 & 103 & -103 \\
-23 & -46 & -92 \\
-23 & 46 & -92 \\
-40 & -120 & -109 \\
-40 & 120 & -109 \\
19 & 76 & 53 \\
19 & -76 & 53 \\
27 & -116 & -78 \\
27 & 116 & -78 \\
0 & 0 & 1
\end{pmatrix}\\ 
& B^T_{_{251}} = \begin{pmatrix}
93 & 0 & 8 & 0 & 115 & 0 & -19 & 0 & 55 & 0 & -1 & 0 \\
0 & 93 & 93 & 101 & 101 & -35 & -35 & -54 & -54 & 1 & 1 & 0 \\
0 & -93 & 93 & -101 & 101 & 35 & -35 & 54 & -54 & -1 & 1 & 0 \\
0 & 79 & -86 & -47 & 102 & 119 & -66 & 102 & 51 & -2 & -1 & 0 \\
0 & -79 & -86 & 47 & 102 & -119 & -66 & -102 & 51 & 2 & -1 & 0 \\
0 & 31 & 94 & 34 & 95 & 70 & 107 & 113 & -46 & 3 & 1 & 0 \\
0 & -31 & 94 & -34 & 95 & -70 & 107 & -113 & -46 & -3 & 1 & 0 \\
0 & -86 & 104 & 24 & 6 & -90 & 103 & -95 & 39 & -4 & -1 & 0 \\
0 & 86 & 104 & -24 & 6 & 90 & 103 & 95 & 39 & 4 & -1 & 0 \\
0 & 119 & 74 & -84 & -67 & 110 & 22 & 101 & -30 & 5 & 1 & 0 \\
0 & -119 & 74 & 84 & -67 & -110 & 22 & -101 & -30 & -5 & 1 & 0 \\
0 & -93 & 0 & -8 & 0 & -115 & 0 & 19 & 0 & -55 & 0 & 1
\end{pmatrix} \\
& A^T_{_{247}} = \begin{pmatrix}
1 & 1 & 1 & 1 & 1 & 1 & 1 & 1 & 1 & 1 & 1 & 0 \\
0 & 1 & -1 & 2 & -2 & 3 & -3 & 4 & -4 & 5 & -5 & 0 \\
0 & 1 & 1 & 4 & 4 & 9 & 9 & 16 & 16 & 25 & 25 & 0 \\
0 & 1 & -1 & 8 & -8 & 27 & -27 & 64 & -64 & -122 & 122 & 0 \\
0 & 1 & 1 & 16 & 16 & 81 & 81 & 9 & 9 & -116 & -116 & 0 \\
0 & 1 & -1 & 32 & -32 & -4 & 4 & 36 & -36 & -86 & 86 & 0 \\
0 & 1 & 1 & 64 & 64 & -12 & -12 & -103 & -103 & 64 & 64 & 0 \\
0 & 1 & -1 & -119 & 119 & -36 & 36 & 82 & -82 & 73 & -73 & 0 \\
0 & 1 & 1 & 9 & 9 & -108 & -108 & 81 & 81 & 118 & 118 & 0 \\
0 & 1 & -1 & 18 & -18 & -77 & 77 & 77 & -77 & 96 & -96 & 1
\end{pmatrix};\ \  
G_{_{247}}  = \begin{pmatrix}
-10 & 0 & 0 \\
74 & 74 & 74 \\
74 & -74 & 74 \\
7 & 14 & 28 \\
7 & -14 & 28 \\
-90 & -23 & -69 \\
-90 & 23 & -69 \\
-20 & -80 & -73 \\
-20 & 80 & -73 \\
-2 & -10 & -50 \\
-2 & 10 & -50 \\
0 & 0 & 1
\end{pmatrix} \\  
&B^T_{_{247}} = \begin{pmatrix}
74 & 0 & -81 & 0 & -12 & 0 & -35 & 0 & 55 & 0 & -1 & 0 \\
0 & 74 & 74 & -7 & -7 & -19 & -19 & -54 & -54 & 1 & 1 & 0 \\
0 & -74 & 74 & 7 & -7 & 19 & -19 & 54 & -54 & -1 & 1 & 0 \\
0 & -37 & 105 & 93 & -77 & 91 & -78 & 102 & 51 & -2 & -1 & 0 \\
0 & 37 & 105 & -93 & -77 & -91 & -78 & -102 & 51 & 2 & -1 & 0 \\
0 & 107 & 118 & -70 & 59 & 98 & 115 & 109 & -46 & 3 & 1 & 0 \\
0 & -107 & 118 & 70 & 59 & -98 & 115 & -109 & -46 & -3 & 1 & 0 \\
0 & 105 & 88 & 104 & 26 & -114 & 95 & -91 & 39 & -4 & -1 & 0 \\
0 & -105 & 88 & -104 & 26 & 114 & 95 & 91 & 39 & 4 & -1 & 0 \\
0 & -84 & 82 & 99 & -79 & -117 & 26 & 97 & -30 & 5 & 1 & 0 \\
0 & 84 & 82 & -99 & -79 & 117 & 26 & -97 & -30 & -5 & 1 & 0 \\
0 & -74 & 0 & 81 & 0 & 12 & 0 & 35 & 0 & -55 & 0 & 1
\end{pmatrix}
\end{align*}
\normalsize 

RNS(253, 251, 247) has the dynamic range of [-7842620, +7842620] being large enough for 8-bit quantized CNN models. The algorithm $F(10 \times 10, 3 \times 3)$ over RNS(253, 251, 247) need 3 element-wise multiplications in 8-bit (accumulation is 32-bit). The implementation can yield up to $2.08\times$ throughput improvement (or Speed-up).

Alternatively, we can compute the Winograd convolution $F(10 \times 10, 3 \times 3)$ over 16-bit RNS(4001, 4331) for instance. 
\begin{align}
{g} \circledast {d} =  
(A^T_{_{4001}} \Big[[G_{_{4001}} {g} G^T_{_{4001}}] \odot [B^{T}_{_{4001}} {d} B_{_{4001}}]\Big] A_{_{4001}},     \nonumber \\
 A^T_{_{4331}} \Big[[G_{_{4331}} {g} G^T_{_{4331}}] \odot [B^{T}_{_{4331}} {d} B_{_{4331}}]\Big] A_{_{4331}})  \label{eq:fast_algorithm_rns_16} 
\end{align}
where the transform matrices are    
\tiny 
\begin{align*}
&A^T_{_{4001}} = \begin{pmatrix}
1 & 1 & 1 & 1 & 1 & 1 & 1 & 1 & 1 & 1 & 1 & 0 \\
0 & 1 & -1 & 2 & -2 & 3 & -3 & 4 & -4 & 5 & -5 & 0 \\
0 & 1 & 1 & 4 & 4 & 9 & 9 & 16 & 16 & 25 & 25 & 0 \\
0 & 1 & -1 & 8 & -8 & 27 & -27 & 64 & -64 & 125 & -125 & 0 \\
0 & 1 & 1 & 16 & 16 & 81 & 81 & 256 & 256 & 625 & 625 & 0 \\
0 & 1 & -1 & 32 & -32 & 243 & -243 & 1024 & -1024 & -876 & 876 & 0 \\
0 & 1 & 1 & 64 & 64 & 729 & 729 & 95 & 95 & -379 & -379 & 0 \\
0 & 1 & -1 & 128 & -128 & -1814 & 1814 & 380 & -380 & -1895 & 1895 & 0 \\
0 & 1 & 1 & 256 & 256 & -1441 & -1441 & 1520 & 1520 & -1473 & -1473 & 0 \\
0 & 1 & -1 & 512 & -512 & -322 & 322 & -1922 & 1922 & 637 & -637 & 1
\end{pmatrix}; \ \  
G_{_{4001}}  = \begin{pmatrix}
222 & 0 & 0 \\
185 & 185 & 185 \\
185 & -185 & 185 \\
-1609 & 783 & 1566 \\
-1609 & -783 & 1566 \\
897 & -1310 & 71 \\
897 & 1310 & 71 \\
1533 & -1870 & 522 \\
1533 & 1870 & 522 \\
-1047 & -1234 & 1832 \\
-1047 & 1234 & 1832 \\
0 & 0 & 1
\end{pmatrix} \\ 
&B^T_{_{4001}} = \begin{pmatrix}
-1604 & 0 & -1071 & 0 & -357 & 0 & -1023 & 0 & 55 & 0 & -1 & 0 \\
0 & -1604 & -1604 & 1326 & 1326 & 969 & 969 & -54 & -54 & 1 & 1 & 0 \\
0 & 1604 & -1604 & -1326 & 1326 & -969 & 969 & 54 & -54 & -1 & 1 & 0 \\
0 & 802 & 401 & 736 & 368 & -1638 & -819 & 102 & 51 & -2 & -1 & 0 \\
0 & -802 & 401 & -736 & 368 & 1638 & -819 & -102 & 51 & 2 & -1 & 0 \\
0 & 799 & 1600 & 1510 & 1837 & 1827 & 609 & -138 & -46 & 3 & 1 & 0 \\
0 & -799 & 1600 & -1510 & 1837 & -1827 & 609 & 138 & -46 & -3 & 1 & 0 \\
0 & 401 & -900 & 1043 & 1261 & -1596 & -399 & 156 & 39 & -4 & -1 & 0 \\
0 & -401 & -900 & -1043 & 1261 & 1596 & -399 & -156 & 39 & 4 & -1 & 0 \\
0 & -1121 & 576 & -99 & -820 & 1365 & 273 & -150 & -30 & 5 & 1 & 0 \\
0 & 1121 & 576 & 99 & -820 & -1365 & 273 & 150 & -30 & -5 & 1 & 0 \\
0 & 1604 & 0 & 1071 & 0 & 357 & 0 & 1023 & 0 & -55 & 0 & 1
\end{pmatrix} \\
&A^T_{_{4331}} = \begin{pmatrix}
1 & 1 & 1 & 1 & 1 & 1 & 1 & 1 & 1 & 1 & 1 & 0 \\
0 & 1 & -1 & 2 & -2 & 3 & -3 & 4 & -4 & 5 & -5 & 0 \\
0 & 1 & 1 & 4 & 4 & 9 & 9 & 16 & 16 & 25 & 25 & 0 \\
0 & 1 & -1 & 8 & -8 & 27 & -27 & 64 & -64 & 125 & -125 & 0 \\
0 & 1 & 1 & 16 & 16 & 81 & 81 & 256 & 256 & 625 & 625 & 0 \\
0 & 1 & -1 & 32 & -32 & 243 & -243 & 1024 & -1024 & -1206 & 1206 & 0 \\
0 & 1 & 1 & 64 & 64 & 729 & 729 & -235 & -235 & -1699 & -1699 & 0 \\
0 & 1 & -1 & 128 & -128 & -2144 & 2144 & -940 & 940 & 167 & -167 & 0 \\
0 & 1 & 1 & 256 & 256 & -2101 & -2101 & 571 & 571 & 835 & 835 & 0 \\
0 & 1 & -1 & 512 & -512 & -1972 & 1972 & -2047 & 2047 & -156 & 156 & 1
\end{pmatrix};\ \  
G_{_{4331}}  = \begin{pmatrix}
1693 & 0 & 0 \\
689 & 689 & 689 \\
689 & -689 & 689 \\
-225 & -450 & -900 \\
-225 & 450 & -900 \\
457 & 1371 & -218 \\
457 & -1371 & -218 \\
1064 & -75 & -300 \\
1064 & 75 & -300 \\
-1626 & 532 & -1671 \\
-1626 & -532 & -1671 \\
0 & 0 & 1
\end{pmatrix} \\ 
& B^T_{_{4331}} = \begin{pmatrix}
1407 & 0 & 579 & 0 & -1017 & 0 & -1023 & 0 & 55 & 0 & -1 & 0 \\
0 & 1407 & 1407 & 1986 & 1986 & 969 & 969 & -54 & -54 & 1 & 1 & 0 \\
0 & -1407 & 1407 & -1986 & 1986 & -969 & 969 & 54 & -54 & -1 & 1 & 0 \\
0 & 1462 & 731 & 76 & 38 & -1638 & -819 & 102 & 51 & -2 & -1 & 0 \\
0 & -1462 & 731 & -76 & 38 & 1638 & -819 & -102 & 51 & 2 & -1 & 0 \\
0 & 469 & 1600 & -2161 & -2164 & 1827 & 609 & -138 & -46 & 3 & 1 & 0 \\
0 & -469 & 1600 & 2161 & -2164 & -1827 & 609 & 138 & -46 & -3 & 1 & 0 \\
0 & 731 & -900 & 713 & 1261 & -1596 & -399 & 156 & 39 & -4 & -1 & 0 \\
0 & -731 & -900 & -713 & 1261 & 1596 & -399 & -156 & 39 & 4 & -1 & 0 \\
0 & -1451 & 576 & 231 & -820 & 1365 & 273 & -150 & -30 & 5 & 1 & 0 \\
0 & 1451 & 576 & -231 & -820 & -1365 & 273 & 150 & -30 & -5 & 1 & 0 \\
0 & -1407 & 0 & -579 & 0 & 1017 & 0 & 1023 & 0 & -55 & 0 & 1
\end{pmatrix}
\end{align*}
\normalsize

The modulus 4001 and 4331 are both coprime to $\frac{1}{\alpha} = 3628800$. The 16-bit RNS has dynamic range $4001 \times 4331 = 17328331$, which allows the convolution output having the maximum magnitude of $\frac{17328331-1}{2}=8664165$. The 16-bit RNS(4001,4331) requires two element-wise multiply, therefore it has arithmetic reduction $3.13\times$, which is better than the $2.08\times$ of 8-bit RNS(253,251,247). But, each element-wise multiplication is of 16-bit op. 

\section{Fast Convolution via integral arithmetic for Convolutional Neural Networks(CNN)} 

Unlike the conventional Winograd algorithm, which could benefit to CNN for both network training and inference, the integer version can only apply to inference of the low-precision (e.g. INT8) quantized CNN models. For a qunatized CNN layer, its major computation is the 2D convolution, $g \circledast x$, between $(R \times R \times C \times K)$ weight tensor ${g}$ and $(B \times W \times H \times C)$ input feature maps ${x}$, where $R \times R$ is the filter size, $C$ is the depth, $K$ is the filter count (or output channels), $B$ is the batch number and $W \times H$ is the dimension of the 2D input plane. 
All elements of $g$ and $x$ are signed integers, e.g. from -128 to 127. Then we can utilize the complexity reduced algorithm, equation \ref{eq:fast_algorithm_rns} or \ref{eq:fast_algorithm_rns_16}  described in section ~\ref{sec:winograd_rns} to compute the integer convolution. 

We can decompose input $x$ into $M\times M$ patches $\{d_i\}$ i.e. $x = \bigoplus\limits_{i}{d_i}$, and apply Winograd algorithm $F(M \times M, R \times R)$ over RNS$(m_0, m_1,..,m_{n-1})$ to each corresponding weight $g$ and patch $d_i$ to compute $g \circledast x$ with the reduced arithmetic as equation ~\ref{eq:direct_sum_winograd}.   
\begin{align} 
    g \circledast x  =  \label{eq:direct_sum_winograd}
    & \bigoplus\limits_{B,K,i} \{ \sum\limits_{C} A^T_{m_0} ((G_{m_0} g^{(C)(K)} G^T_{m_0}) \odot (B^T_{m_0} d^{(C)(B)}_i B_{m_0})) A_{m_0}, \\ \nonumber 
    & \sum\limits_{C} A^T_{m_1} ((G_{m_1} g^{(C)(K)} G^T_{m_1}) \odot (B^T_{m_1} d^{(C)(B)}_i B_{m_1})) A_{m_1}, .., \\ 
    & \sum\limits_{C} A^T_{m_{n-1}} ((G_{m_{n-1}} g^{(C)(K)} G^T_{m_{n-1}}) \odot (B^T_{m_{n-1}} d^{(C)(B)}_i B_{m_{n-1}})) A_{m_{n-1}} \}  \nonumber \\
= \label{eq:direct_sum_2}  
\{ & A^T_{m_0}(\bigoplus\limits_{B,K,i}(\sum\limits_{C}  ((G_{m_0} g^{(C)(K)} G^T_{m_0}) \odot (B^T_{m_0} d^{(C)(B)}_i B_{m_0})))) A_{m_0}, \\ \nonumber
    & A^T_{m_1} (\bigoplus\limits_{B,K,i}( \sum\limits_{C} ((G_{m_1} g^{(C)(K)} G^T_{m_1}) \odot (B^T_{m_1} d^{(C)(B)}_i B_{m_1})))) A_{m_1}, \nonumber \\ & \ \ \ ... \ , \nonumber \\ 
    & A^T_{m_{n-1}} (\bigoplus\limits_{B,K,i}( \sum\limits_{C}  ((G_{m_{n-1}} g^{(C)(K)} G^T_{m_{n-1}}) \odot (B^T_{m_{n-1}} d^{(C)(B)}_i B_{m_{n-1}})))) A_{m_{n-1}} \}  \nonumber
\end{align}
\normalsize

In eq. ~\ref{eq:direct_sum_2}, the forward Winograd Transform of filter e.g. $G_{m_0} w^{(C)(K)} G^T_{m_0}$ can be pre-calculated. The forward transform of input e.g. $ B^T_{m_0} x^{(C)(B)}_i B_{m_0} $ is shared or reused across $K$ filters, therefore their computation cost got amortized by factor $K$. The backward transform was performed after the reduction across depth $C$ due to linearity of transform, so the backward transform was amortized by factor of $C$. 

The point-wise multiply terms in eq. ~\ref{eq:direct_sum_2}, for instance, 
\begin{equation} \label{eq:gemm}
\bigoplus\limits_{B,K,i} (\sum\limits_{C} ((G_{m_0} g^{(C)(K)} G^T_{m_0}) \odot (B^T_{m_0} d^{(C)(B)}_i B_{m_0}))) \pmod{m_0}
\end{equation}
(eq. ~\ref{eq:gemm}) is a matrix multiply (GEMM) function essentially followed by a modulo operation, which can be executed by existing highly optimized GEMM library, such as gemmlowp \footnote{\url{https://github.com/google/gemmlowp}} or accelerator. Notably, we can perform the modulo operation after the GEMM to reduce its overhead. In the final step after the backward transform, we convert the $g \circledast x$ from the RNS presentation to the standard format using MRC or CRT. 

\section{Performance Analysis} \label{sec:performance}

The performance of RNS based Winograd convolution depends on the transformation and filter size i.e. $N=M+R-1$ and $R$ respectively. When computation is carried out in RNS and the cost of Winograd transformation and MRC are amortized due to reuse, the theoretical arithmetic reduction is given by 
\begin{equation} \label{eq:speedup_rns} \nonumber 
\frac{M^2 R^2}{N^2} \times \frac{1}{n}
\end{equation}
where $n$ is the modulus number of the RNS. Table ~\ref{tab:speedup} contains the complexity reduction for different algorithms, $F(M \times M, R \times R)$. The Winograd algorithm has better complexity reduction for large values of $M$ and achieves more benefit for $5 \times 5 $ filters than the $3 \times 3$. Moreover, 2-residue RNS, such as RNS(4001,4331), has more arithmetic reduction than 3-residue case. For example, $F(12 \times 12, 5 \times 5)$ over RNS(4001,4331) generates 7.03$\times$ reduction vs $4.69 \times $ over RNS(251,241,239). However, 2-residue RNS(4001,4331) requires 16-bit GEMM operation, which will be less efficient than the 8-bit case regarding throughput and power consumption.       

\begin{table*}[hbt] 
\caption{Complexity reduction of Winograd convolution in RNS. }
\label{tab:speedup}
\scriptsize
\centering
\begin{tabular}{l | c  |  c}
\toprule
Algorithms & \multicolumn{2}{c}{Arithmetic Complexity Reduction}  \\
\cmidrule{2-3}  
$F$($M\times M$,$R \times R$) & \ \    RNS(4001,4331) \ \ \    &
RNS(251,241,239) \\ 
\midrule
$F$($2\times2$, $3\times 3$) &  1.125$\times$ & \st{0.75$\times$} \\
$F$($4\times4$, $3\times 3$) &  2.00$\times$  & 1.33$\times$ \\
$F$($6\times6$, $3\times 3$) &  2.53$\times$  & 1.69$\times$ \\
$F$($8\times8$, $3\times 3$) &  2.88$\times$  & 1.92$\times$ \\
$F$($8\times8$, $5\times 5$) &  5.56$\times$  & 3.70$\times$ \\
$F$($9\times9$, $3\times 3$) &  3.01$\times$ & 2.01$\times$ \\
$F$($9\times9$, $5\times 5$) &  5.99$\times$ & 3.99$\times$ \\
$F$($10\times10$, $3\times 3$) & 3.13$\times$ & 2.08$\times$ \\
$F$($10\times10$, $5\times 5$) &  6.38$\times$ & 4.25$\times$ \\
$F$($11\times11$, $3\times 3$) &  3.22$\times$ & 2.14$\times$ \\
$F$($11\times11$, $5\times 5$) &  6.72$\times$ & 4.48$\times$ \\
$F$($12\times12$, $3\times 3$) &  3.31$\times$  & 2.20$\times$ \\
$F$($12\times12$, $5\times 5$) &  7.03$\times$  & 4.69$\times$ \\
$F$($14\times14$, $3\times 3$) &  3.45$\times$  & 2.30$\times$ \\
\bottomrule 
\end{tabular}
\end{table*}

Our RNS approach is in favor of large transformation, such as $10 \times 10$ to $16 \times 16 $ etc. since the numerical issue is mitigated by using RNS. However, the computation cost of Winograd transform, both forward and backward ones, will be higher than using small transformation.

The critical path of the computation is the element-wise multiplication, which is low-precision GEMM operations. Table ~\ref{tab:gemm} shows the throughput in GOPS(Giga $(10^9)$ Operations Per-Second) of 8-bit and 16-bit GEMM measured on a single core of Arm Cortex-A73 CPU for variety of size and shape. 

For a given hardware e.g. CPU, GPU or accelerator we can determine the optimal implementation based on table ~\ref{tab:speedup} and the corresponding GEMM performance. For example, targeting Arm Cortex-A73 CPU used in the benchmark, if we choose RNS(4001,4331) to compute the convolution using $F(12 \times 12, 5 \times 5)$ with $1024 \times 1024 \times 1024$ GEMM, it will have a theoretical speed-up up to $\frac{7.03 \times 9.58}{14.6} = 4.6\times $ to the Im2col+INT16GEMM baseline, while the improvement of RNS(253,251,247) is about $\frac{4.69 \times 14.6}{14.6} = 4.69\times$ over Im2col+INT8GEMM. So, RNS(251,241,239) and RNS(4001,4331) happen to deliver roughly the same improvement with the benchmark program on the Cortex-A73 CPU specifically, but in general 8-bit implementation RNS(251, 241, 239) will consume less power since it uses 8-bit arithmetic. Other hardware, for example Nvidia's RTX2020Ti GPU with up to 215 TOPS of INT8 ops \footnote{\url{https://devblogs.nvidia.com/nvidia-turing-architecture-in-depth/}}, could potentially gain up to a factor of $2.30\times$ or $4.69\times$ performance boost for $3 \times 3$ or $5 \times 5 $ filters respectively through $16 \times 16$ RNS-Winograd transformation.

\section{Experiments}

To validate the proposal, the RNS based Winograd convolution algorithm was implemented in a highly optimized kernel in C on Ubuntu Linux. The program takes advantage of ILP (vector units) to boost the throughput of Winograd transforms, MRC and GEMM functions. 


\begin{table}[bt] 
\caption{Throughput (GOPS) of 8-bit, 16-bit and 32-bit GEMM (32-bit output) on 1 CPU of Arm Cortex-A73. e.g. $1024 \times 64\times 1024$ GEMM indicates the matrix multiply of $1024 \times 64 $ by $64 \times 1024$. }
\label{tab:gemm}
\scriptsize
\centering
\begin{tabular}{c  c  c}
\toprule
GEMM &\ \  8-bit GOPS\ \  & \ \  16-bit GOPS\ \  \\
\midrule  
1024$\times$64$\times$1024   & 11.1  &  8.46  \\
1024$\times$128$\times$1024  & 13.3  &  10.1 \\
1024$\times$256$\times$1024  & 14.8  &  10.9  \\

256$\times$256$\times$256    & 14.5  &  11.1  \\
512$\times$512$\times$512    & 15.4  &  11.2  \\

1024$\times$1024$\times$1024 & 14.6  &  9.58  \\
2048$\times$2048$\times$2048 & 14.2  &  11.2  \\
4096$\times$4906$\times$4096 & 14.5  &  9.83  \\
\bottomrule 
\end{tabular}
\end{table}

\begin{table*}[!hbt]
\caption{Inference performance of 8-bit activation and 8-bit weight quantized CNN layers of VGG16 with Winograd algorithm $F(14 \times 14, 3 \times 3)$ over RNS(251,241,239) and RNS(4001,4331) on Arm Cortex-A73, having 71.4\% top-1 prediction accuracy with ImageNet dataset. The corresponding transforms are in the supplementary materials. The speed-up of RNS(251,241,239) and RNS(4001,4331) are the runtime improvement relative to the standard INT8 and INT16 Im2col+GEMM convolution baselines respectively.}
\scriptsize
\label{tab:vgg16}
\centering
\begin{tabular}{c|c|c|c c}
\toprule
\ VGG16 \ \ & {conv2d op}           & Winograd  &  \multicolumn{2}{c} {Speed-up}  \\
model & (int8) x (int8)       & Algorithm &  \ RNS(251,241,239) & \ RNS(4001,4331)\ \\
\midrule
conv1\_1 & $(224,224,3)\times(3,3,3,64)$                        &   -$^\dagger$    & 1$\times$  &    1$\times$      \\
conv1\_2      & $(224,224,3)\times(3,3,3,64)$                   &  $F(14 \times 14 ,3 \times 3)$      &  1.86$\times$   &  2.05$\times$  \\
conv2\_1      & $(112,224,64)\times(3,3,64,64)$                 &  $F(14 \times 14 ,3 \times 3)$      &  1.97$\times$   &  2.13$\times$ \\
conv2\_2      & $(112,112,64)\times(3,3,64,128)$                &  $F(14 \times 14 ,3 \times 3)$      &  2.07$\times$   &  2.25$\times$  \\
conv3\_1      & $(56,56,128)\times(3,3,128,128)$                &  $F(14 \times 14 ,3 \times 3)$      &  2.14$\times$   &  2.33$\times$  \\
conv3\_2      & $(56,56,128)\times(3,3,128,256)$                &  $F(14 \times 14 ,3 \times 3)$      &  2.15$\times$   &  2.37$\times$  \\
conv3\_3      & $(56,56,256)\times(3,3,256,256)$                &  $F(14 \times 14 ,3 \times 3)$      &  2.16$\times$   &  2.35$\times$  \\
conv4\_1      & $(28,28,256)\times(3,3,256,512)$                &  $F(14 \times 14 ,3 \times 3)$      &  2.21$\times$   &  2.40$\times$   \\ 
conv4\_2      & $(28,28,512)\times(3,3,512,512)$                &  $F(14 \times 14 ,3 \times 3)$      &  2.25$\times$   &  2.37$\times$  \\ 
conv4\_3      & $(28,28,512)\times(3,3,512,512)$                &  $F(14 \times 14 ,3 \times 3)$      &  2.27$\times$   &  2.39$\times$   \\ 
conv5\_1      & $(14,14,512)\times(3,3,512,512)$                &  $F(14 \times 14 ,3 \times 3)$      &  2.21$\times$   &  2.44$\times$  \\ 
conv5\_2      & $(14,14,512)\times(3,3,512,512)$                &  $F(14 \times 14 ,3 \times 3)$      &  2.24$\times$   &  2.39$\times$   \\ 
conv5\_3      & \ $(14,14,512)\times(3,3,512,512)$      \      & \  $F(14 \times 14 ,3 \times 3) \ $  &  2.22$\times$   &  2.43$\times$  \\ 
\midrule
\multicolumn{3}{l }{average}      &   2.02$\times$ &  2.20$\times$    \\
\bottomrule
\end{tabular} \\
$^\dagger$ Fallback to the baseline. 
\end{table*}

The 2D convolution of 8-bit quantized (for both weight and activation) VGG16 network was benchmarked using the RNS based Winograd algorithm implemented on Arm Cortex-A73 CPU. The convolution output of all CNN layers are within the range of $[-3.0\times 10^5, 3.0\times10^5]$ measured from validation images of ImageNet dataset. We used RNS(251,241,239) and RNS(4001,4331), which have the large enough dynamic ranges, $[-7228674, 7228674]$ and $[-8664165, 8664165]$ respectively to guarantee the correctness of the computation. 

Using algorithm $F(14 \times 14, 3 \times 3)$, the performance improvement or speed-up over the Im2col+INT8/16 GEMM baselines for both 8-bit and 16-bit RNS are listed in table \ref{tab:vgg16}. The overall convolution computation latency reduction is $\textbf{2.02}\times$ for 8-bit RNS(251,241,239) or $\textbf{2.20}\times$ for 16-bit RNS(4001,4331). On average, the execution overheads, measured in time, of the 8-bit RNS(251, 241, 239) are 7.9\% for the forward Winograd Transform of input feature maps, 9.2\% for the backward Winograd transform of output, and 1.1\% for MRC while for the 16-bit RNS(4001, 4331), the corresponding overheads are 9.4\%, 10.2\%, and 1.3\% respectively. Table \ref{tab:other_models} provides extra experimental results for 8-bit ResNet50-v1 and Inception v1 and v3 models using INT8 arithmetic ops. Notably, the Inception-v3 contains three $5 \times 5$ convolutional layers,  
(1) Mixed\_5/Branch\_1/ Conv2d\_0b\_5x5, 
(2) Mixed\_5c/Branch\_1/Conv\_1\_0c\_5x5 and 
(3) Mixed\_ 5d/Branch\_1/Conv2d\_0b\_5x5 
with $(5 \times 5 \times 48 \times 64)$ kernels. The average speed-up for the $5 \times 5$ layers are $2.31\times$ with 8-bit 3-residue RNS. 

\begin{table*}[!htb]
\caption{Inference performance improvement over the Im2col+INT8GEMM baseline of CNN layers for 8-bit quantized ResNet50-v1, Inception v1 and v3 models with ImageNet dataset, using 8-bit RNS(251,241,239).}
\scriptsize
\label{tab:other_models}
\centering
\begin{tabular}{c  c  c  c}
\toprule
\ Models & \ \ \ Bits of weight/input \ \ \  & \ \ \ Top-1 Acc.(\%) \ \ \ & \ \ \ Speed-up of CNN layers$^\dagger$ \\
\midrule
ResNet50-v1 & 8/8 & 75.1 & 1.76$\times$ \\
Inception-v1 & 8/8 & 70.1 & 1.82$\times$ \\
Inception-v3 & 8/8 & 77.5 & 1.35$\times$ \\
\bottomrule
\end{tabular} \\
$^\dagger$ Not include the CNN layers with the stride $\ge$ 2. 
\end{table*}

\section{Conclusions}

We proposed a Residue Number System (RNS) based fast integral Winograd convolution that overcomes the computational instability of the conventional Winograd algorithm. The method enables the execution of the Winograd algorithm using low cost, low precision arithmetic operations (e.g. INT8 MAC) for inference of existing quantized CNN networks. The convolution outputs are precise, which means there is no prediction accuracy degradation with the RNS-based Winograd convolution scheme we have presented.

Our RNS-based approach can benefit the common hardware platforms, including CPU, GPU, and hardware accelerators, which can deliver high throughput, low cost integer MAC operations. The theoretical performance improvement of 8-bit quantized CNN layers can be up to ${2.3}\times$ and ${4.6}\times$ over 8-bit 3-residue RNS for $3\times 3 $ and $5\times 5$ CNN layers respectively using up to $16\times16$ transformation. 

The experiment showed, on average, the new proposal improved the runtime performance of $3\times 3$ INT8 CNN layers by ${2.02}\times$ using power efficient 8-bit arithmetic and ${2.20}\times$ for 16-bit arithmetic over the standard Im2col + INT8 and INT16 GEMM baseline performances respectively measured on an Arm Cortex-A73 mobile CPU using the 8-bit quantized VGG16 model,  including the computation overheads such as Winograd transforms over RNS, modulo, and MRC operations etc. The new proposal  achieved higher improvement e.g. $2.31\times$ for the CNN layers with larger filter size i.e. $5 \times 5$ in Inception-v3.

Although it is possible to increase the transformation size (i.e. $>16\times16$), to further boost arithmetic reduction, the transformation cost increases roughly linearly, therefore it is a reasonable trade-off to choose transformation size from 8 to 16.

\newpage 
%
%
\bibliographystyle{splncs04}
\bibliography{egbib}

\end{document}